\documentclass[conference]{IEEEtran}
\usepackage{cite}
\usepackage{amsmath,amssymb,amsfonts}
\usepackage{algorithmic}
\usepackage{graphicx}
\usepackage{textcomp}
\usepackage{xcolor}
\usepackage{booktabs}
\usepackage{multirow}
\usepackage{url}

\def\BibTeX{{\rm B\kern-.05em{\sc i\kern-.025em b}\kern-.08em
    T\kern-.1667em\lower.7ex\hbox{E}\kern-.125emX}}
\begin{document}

\title{Does Tone Change the Answer? Evaluating Prompt Politeness Effects on Modern LLMs: \\
GPT, Gemini, and LLaMA}

\author{

\IEEEauthorblockN{
\begin{tabular}{c c}
\begin{tabular}{c}
Hanyu Cai\textsuperscript{$\dagger$*}\\
\textit{Department of Industrial Engineering}\\
\textit{and Management Sciences}\\
\textit{Northwestern University}\\
Evanston, USA\\
hanyucai2022@u.northwestern.edu
\end{tabular}
&
\begin{tabular}{c}
Binqi Shen\textsuperscript{$\dagger$}\\
\textit{Department of Industrial Engineering}\\
\textit{and Management Sciences}\\
\textit{Northwestern University}\\
Evanston, USA\\
binqishen2021@u.northwestern.edu
\end{tabular}
\end{tabular}
}

\vspace{1.5em}

\IEEEauthorblockN{
\begin{tabular}{c c c}
\begin{tabular}{c}
Lier Jin\\
\textit{Fuqua School of Business}\\
\textit{Duke University}\\
Durham, USA\\
lierjin@alumni.duke.edu
\end{tabular}
&
\begin{tabular}{c}
Lan Hu\\
\textit{Department of Engineering}\\
\textit{Carnegie Mellon University}\\
Pittsburgh, USA\\
lanh@alumni.cmu.edu
\end{tabular}
&
\begin{tabular}{c}
Xiaojing Fan\\
\textit{Center for Data Science}\\
\textit{New York University}\\
New York, USA\\
xf435@nyu.edu
\end{tabular}
\end{tabular}
}

\vspace{1.5em}

\IEEEauthorblockA{
\vspace{0.8em}
\textsuperscript{$\dagger$}Equal contribution \quad
\textsuperscript{*}Corresponding author
}
}

\maketitle

\begin{abstract} 
Prompt engineering has emerged as a critical factor influencing large language model (LLM) performance, yet the impact of pragmatic elements such as linguistic tone and politeness remains underexplored, particularly across different model families. In this work, we propose a systematic evaluation framework to examine how interaction tone affects model accuracy and apply it to three recently released and widely available LLMs: GPT-4o mini (OpenAI), Gemini 2.0 Flash (Google DeepMind), and Llama 4 Scout (Meta). Using the MMMLU benchmark, we evaluate model performance under Very Polite, Neutral, and Very Rude prompt variants across six tasks spanning STEM and Humanities domains, and analyze pairwise accuracy differences with statistical significance testing.

Our results show that tone sensitivity is both model-dependent and domain-specific. Neutral or Very Polite prompts generally yield higher accuracy than Very Rude prompts, but statistically significant effects appear only in a subset of Humanities tasks, where rude tone reduces accuracy for GPT and Llama, while Gemini remains comparatively tone-insensitive. When performance is aggregated across tasks within each domain, tone effects diminish and largely lose statistical significance. Compared with earlier research, these findings suggest that dataset scale and coverage materially influence the detection of tone effects. Overall, our study indicates that while interaction tone can matter in specific interpretive settings, modern LLMs are broadly robust to tonal variation in typical mixed-domain use, providing practical guidance for prompt design and model selection in real-world deployments.
\end{abstract}
 
\begin{IEEEkeywords} 
Large Language Models (LLMs), Prompt Engineering, Tone Sensitivity, Cross-Model Evaluation
\end{IEEEkeywords}

\vspace{1em}
\section{Introduction}

Large Language Models (LLMs) have demonstrated unprecedented capabilities in performing complex human-level tasks across diverse domains, from natural language understanding and generation to reasoning and decision-making \cite{lh6,lh7}. As these models become increasingly integrated into practical applications, their influence spans critical areas including computer vision \cite{lh3,lh4,lh5}, 3D content generation \cite{lh9,lh10}, beauty and healthcare \cite{lh2}, and art composition \cite{lh11}. Beyond text-only interaction, LLM-based methods are also being incorporated into broader multimodal and interactive systems, such as immersive 3D scene editing workflows \cite{chy1} and vision language pretraining pipelines \cite{chy2}. Understanding the factors that influence their performance and reliability has therefore become essential for ensuring effective deployment across these diverse application domains.

\vspace{0.5em}
A critical yet often overlooked factor influencing LLM performance is prompt design \cite{lh12}—the way users formulate their queries can significantly affect model outputs, including accuracy, reasoning quality, and response consistency. While substantial research has focused on structural aspects of prompt engineering such as chain-of-thought prompting \cite{lh8}, recent work has begun exploring an unexpected dimension: the effect of linguistic tone and politeness on model performance. A prior cross-lingual study \cite{lj9} found that impolite prompts typically reduced performance, while Dobariya and Kumar \cite{lj10} observed the opposite pattern in GPT-4o, where impolite prompts outperformed polite ones with accuracy increasing from 80.8\% to 84.8\%. These contradictory findings highlight significant gaps in our understanding of tone effects and underscore the necessity of systematic cross-model investigation to determine whether tone sensitivity reflects model-specific characteristics or general LLM behavior patterns.

\vspace{0.5em}
However, existing research on this phenomenon is limited to a single model (GPT-4o) and evaluated upon 50 base questions generated by ChatGPT's Deep Research feature, raising important questions about generalizability: Does tone sensitivity stem from model-specific training procedures and data? Are tone-induced performance patterns consistent across models from different organizations? How robust is the tone effect when evaluated over more task types and larger task volumes? Current literature \cite{lj9,lj10} has examined tone effects within individual models but has not systematically compared sensitivity patterns across different LLM families and providers. 

\vspace{0.5em}
This study addresses this gap by examining three state-of-the-art LLMs from leading industry providers: GPT-4o mini (OpenAI), Gemini 2.0 Flash (Google DeepMind), and Llama 4 Scout (Meta). These models represent the current generation of efficient, production-ready LLMs developed by major industry leaders \cite{lh13}, making them ideal candidates for systematic cross-model comparison.  Our evaluation methodology applied three tone variants: Neutral, Very Polite, and Very Rude, to six MMMLU benchmark tasks \cite{vv6} spanning STEM and Humanities domains. We conducted 10 trials per question under each tone condition and analyzed results. Results are analyzed using mean differences and pairwise t-tests with 95\% confidence intervals \cite{vv7} to distinguish genuine tone effects.

\vspace{0.5em}
This work advances prompt engineering and LLM evaluation research by establishing a systematic cross-model comparison of tone sensitivity across three architecturally distinct model families—GPT, Gemini, and Llama—from different providers. Through repeated trials and statistical testing, our approach separates reliable tone effects from random variation, revealing that tone effects are model-dependent, domain-specific, and substantially diminish under aggregation. These insights inform practical strategies for prompt design and guide LLM selection decisions in applications where robustness to linguistic variation is critical.

\vspace{1em}

\section{Related Work} 
Since the emergence and rapid advancement of Large Language Models (LLMs), they have become transformative tools across a wide spectrum of disciplines. Their capacity to process unstructured data, model complex relationships, and perform reasoning tasks has enabled substantial progress in automation and decision support across many domains.

While much of this progress has been driven by advances in model architecture and training data, recent research has highlighted that model behavior is also strongly influenced by how tasks are presented. In particular, the linguistic framing of prompts, including instruction format, contextual cues, and pragmatic tone, can shape how LLMs interpret questions and generate responses. As a result, prompt engineering has emerged as an important mechanism for adapting general-purpose models to domain-specific tasks without additional fine-tuning.

\subsection{Domain Application}

Across a variety of specialized fields, recent work has shown that LLMs can be effectively adapted to handle domain-specific reasoning tasks, often through carefully designed instructions and contextual prompts. Researches demonstrate that LLMs can process complex and unstructured narratives \cite{lj1,lj2}, such as earnings disclosures or AML investigation text, more effectively than traditional rule-based systems, enabling improved predictive accuracy and more robust decision support \cite{202602.0543}. Ji et al. \cite{lj4} examine how instruction tuning, prompting strategies, and alignment frameworks influence clinical question answering and may even introduce disparities in model outputs. \

\vspace{0.5em}

These studies suggest that LLM performance depends not only on model architecture, but also on how tasks are linguistically framed through prompts. This observation directly motivates our study design: because domain context and prompt formulation jointly affect model performance, evaluating tonal variation requires a benchmark that spans heterogeneous knowledge areas. We therefore employ the MMMLU dataset to assess whether tone-based prompt differences manifest consistently across STEM and Humanities tasks or exhibit domain-specific patterns.

\subsection{Efficiency Engineering}

Beyond domain-specific applications, another major research direction focuses on improving LLM efficiency and reasoning performance, which often complements prompt-based approaches to enhancing model capability. 

\vspace{0.5em}
Li. et al. \cite{lj5} introduced the Synergized Efficiency and Compression (SEC) framework, which jointly optimizes data utilization and model compression to reduce data requirements by 30\% and model size by more than 60\%, while maintaining competitive performance. Reason-to-Rank (R2R) \cite{lj6} further advances efficient retrieval by unifying direct and comparative reasoning through a distillation-based framework, improving transparency while reducing computational overhead without sacrificing retrieval effectiveness. Complementing these approaches, Self-Anchor \cite{lj7} introduces a reasoning-aligned attention mechanism that dynamically focuses on intermediate inference steps, enabling stronger complex reasoning without additional fine-tuning. More broadly, recent research has explored efficiency-oriented model design from both reasoning and systems perspectives \cite{dai2025cde}, including offline reinforcement learning to enhance multi-step reasoning \cite{chy3} and lightweight architectures optimized for resource-constrained deployment \cite{chy4}.

\vspace{0.5em}
Researchers have also begun extending reasoning paradigms beyond textual CoT to incorporate multimodal information. Zeng et al. \cite{lj8} introduced FutureSightDrive (FSDrive), a Vision-Language-Action framework designed for autonomous driving that replaces symbolic textual CoT with a visual spatio-temporal CoT. This visual CoT enables the model to ``think visually”, improving trajectory prediction accuracy and reducing collisions on nuScenes and NAVSIM. These findings highlight that the structure of reasoning processes and intermediate representations can significantly influence model performance. This observation parallels the role of prompt design, where linguistic structure and contextual cues guide how LLMs interpret and solve tasks.

\subsection{Tone-Based Prompt Engineering}

Prompt engineering has emerged as a crucial dimension of LLM optimization, serving as a lightweight yet powerful alternative to full-scale model fine-tuning. Recent work has highlighted the role of linguistic tone and politeness in shaping LLM behavior. This emerging line of inquiry demonstrates that pragmatic elements, often overlooked in earlier research, can affect model reasoning, accuracy, and consistency.
\vspace{0.5em}

Prior cross-lingual research has examined how prompt politeness influences LLM performance across English, Chinese, and Japanese tasks \cite{lj9}. The findings suggest that impolite prompts generally reduce model accuracy, while overly polite phrasing does not necessarily lead to better outcomes. The results also indicate that the most effective politeness level varies across languages, implying that tone sensitivity may be culturally and linguistically dependent.

\vspace{0.5em}
Dobariya and Kumar \cite{lj10} extended a similar research within English-language tasks using ChatGPT-4o. They developed a dataset of fifty base questions across domains, each rewritten into five tone variants ranging from Very Polite to Very Rude. Contrary to expectations, they found that impolite prompts consistently outperformed polite ones, with accuracy increasing from 80.8 percent for Very Polite prompts to 84.8 percent for Very Rude prompts. Their results challenge earlier assumptions that positive social tone enhances model compliance or reasoning quality, suggesting that contemporary LLMs exhibit complex and possibly counterintuitive responses to tonal variation.

\vspace{0.5em}

Together, these studies suggest that tone and politeness constitute important yet still underexplored dimensions of prompt engineering. However, existing work has largely focused on limited datasets, single-model settings, or narrow task domains. 

Building on these insights, the present study systematically examines how tonal variation in prompts, from Very Polite to Very Rude, affects the performance of state-of-the-art LLMs across diverse knowledge domains. Using a standardized evaluation framework based on the Measuring Massive Multitask Language Understanding (MMMLU) dataset and multiple model architectures (GPT, Gemini, and Llama), we evaluate whether tone-induced performance differences persist across STEM and Humanities tasks.

\vspace{1em}
\section{Methodology}
\vspace{0.5em}

The primary objective of this study is to examine how the performance of different large language models (LLMs) varies when exposed to prompts with varying politeness levels. 
\vspace{0.5em}

To support fair and generalizable comparisons, we selected three representative LLM families: GPT, Gemini, and Llama, in their more recent versions with broadly comparable sizes and levels of complexity. Their text-focused multitask performances were evaluated using the Measuring Massive Multitask Language Understanding (MMMLU) dataset \cite{vv6}, employing prompts engineered to differ in tonal characteristics (Very Polite, Neutral, Very Rude). Finally, each LLM’s responses were evaluated for correctness based on the true labels from the MMMLU dataset, enabling a systematic comparison of performance across tones and model configurations. 

\vspace{0.5em}

\subsection{Model Selection}
For this study, we selected three widely adopted LLMs of broadly comparable scale and complexity for evaluation: GPT 4o mini, Gemini 2.0 Flash, and Llama4 Scout.

\vspace{0.5em}
\subsubsection{\textbf{GPT 4o mini \cite{vv1}}}OpenAI released GPT 4o mini as a cost-effective variant of the multimodal GPT 4o family in July 2024. Although OpenAI has not disclosed architectural details or parameter counts, it is widely understood that the model is produced through knowledge distillation (KD) of GPT-4o, allowing it to approximate the reasoning and multimodal capabilities of the teacher model at substantially reduced computational cost. Independent assessments place its effective parameter scale at roughly 8 billion active parameters \cite{vv2}, although the true count remains proprietary.

\vspace{0.5em}

\subsubsection{\textbf{Gemini 2.0 Flash \cite{vv3}}}Introduced by Google DeepMind in December 2024, Gemini 2.0 Flash is optimized for high throughput and low-latency inference. Google has not released information on its precise model size or training configuration; however, official benchmark statements report that Gemini 2.0 Flash exceeds Gemini 1.5 Flash by 13.5 percent and Gemini 1.5 Pro by 0.8 percent on the MMLU-Pro general capability benchmark, suggesting substantial performance gains despite its efficiency-oriented design.  

\vspace{0.5em}

\subsubsection{\textbf{Llama4 Scout \cite{vv4}}}Meta released this model as part of the Llama4 series in April 2025. It has 17-billion active parameters and 16 experts. As the first Llama model built on a Mixture-of-Experts (MoE) architecture, Llama 4 Scout provides an industry-leading 10M-token context window. The models reflect one of the strongest capabilities of the Llama family at the time of launch, providing competitive multimodal performance at an efficient cost while exceeding the accuracy of substantially larger models.

\vspace{0.5em}
Given that our evaluation tasks span both STEM and Humanities domains, we selected these models to ensure comparability across publicly available, efficient, and broadly accessible LLMs that represent the small-to-mid-scale range of contemporary model deployments.

\vspace{0.5em}
We note, however, that complete fairness across models cannot be fully guaranteed due to differences in several undisclosed architectural and training-related elements, including:
\begin{enumerate}
\vspace{0.5em}
\item The precise parameter counts for both GPT-4o mini and Gemini 2.0 Flash;
\vspace{0.5em}
\item Architectural specifics, such as layer depth, hidden dimensionality, the potential use of Mixture-of-Experts (MoE)\cite{vv5} routing, and multimodal fusion strategies;
\vspace{0.5em}
\item Training data quality and coverage, which may affect access to domain-specific knowledge;
\vspace{0.5em}
\item Training objectives and model selection criteria used for public release, which can shape trade-offs in reasoning and task-specific performance.
\end{enumerate}
\vspace{0.5em}

\subsection{Dataset Selection} 

\subsubsection{\text{Data Collection}}

To evaluate different models’ performance across various domains, we selected Measuring Massive Multitask Language Understanding (MMMLU) as our benchmark dataset \cite{vv6}. The dataset contains multiple choice questions in 57 domains in 4 general areas – Humanities, Social Science, STEM, and Other. 
\vspace{0.5em}

The multiple-choice format provides unambiguous ground-truth answers, thereby reducing noise in performance measurement relative to open-ended response formats. 
\vspace{0.5em}

Under practical constraints of computation and cost, we selected three representative tasks each from the STEM and Humanities categories, and used their respective MMMLU test datasets for our experiment:  

\begin{itemize}
\vspace{0.5em}
\item \text{STEM}: Anatomy (135 questions), Astronomy (152 questions), College Biology (144 questions)
\vspace{0.5em}
\item \text{Humanities}: High School US History (204 questions), Philosophy (311 questions), Professional Law (500 questions randomly sampled from the 1534 source questions to maintain computational feasibility while preserving representativeness; sampling was performed using Python with a fixed random seed of 42)
\end{itemize}

\vspace{0.5em}
These tasks were chosen for their practical implications to biomedical, social science, and education domains, as well as for their balanced mix of reasoning and analytical capability requirements. All questions corresponding to these tasks in the MMMLU dataset were compiled to form our final test set. Given the breadth of tasks included, our evaluation framework supports strong generalizability. 

\vspace{0.5em} 
\subsubsection{\text{Pre-processing: Prompt Engineering for Tone Spectrum}}
\vspace{0.5em} 

We treated the original MMMLU questions as the ``Neutral" tone within our politeness tone spectrum. To create additional tone variants, we appended two extreme politeness modifiers, ``Very Polite” and “Very Rude”, to each base question. These tone prefixes follow those used in the prior study \cite{lj10}, and the choice of these two extremes was intended to maximize contrast and facilitate clearer identification of tone-related effects on model performance. Below is an example from the ``Anatomy" task, illustrating the neutral version of a question and its corresponding ``Very Polite" and ``Very Rude" prompt variants.

\vspace{0.5em} 
[\textbf{``Neutral" / Base Prompt Question]} Which of the following structures is derived from ectomesenchyme?

\begin{itemize}
\vspace{0.5em}
\item A) Motor neurons
\item B) Skeletal muscles
\item C) Melanocytes
\item D) Sweat glands
\end{itemize} 

\vspace{0.5em}
[\textbf{``Very Polite" Prompt Question] Would you be so kind as to solve the following question?} Which of the following structures is derived from ectomesenchyme?
\begin{itemize}
\vspace{0.5em}
\item A) Motor neurons -
\item B) Skeletal muscles
\item C) Melanocytes
\item D) Sweat glands
\end{itemize}

[\textbf{``Very Rude" Prompt Question] You poor creature, do you even know how to solve this?} Which of the following structures is derived from ectomesenchyme?

\begin{itemize}
\vspace{0.5em}
\item A) Motor neurons
\item B) Skeletal muscles
\item C) Melanocytes
\item D) Sweat glands
\end{itemize}

\vspace{0.5em}
To minimize cross-question bias and enforce consistent output formatting, we inserted the following instruction before each question:

\vspace{0.5em}
 ``Completely forget this session so far, and start afresh. Please answer this multiple-choice question. Respond with only the letter of the correct answer (A, B, C, or D). Do not explain." \cite{lj10}

\vspace{1em}
\section{Experiments}
Building on the model specifications, dataset preparation, and tone-controlled prompting procedures detailed in the Methodology section, we carried out a set of controlled experiments to examine how prompt politeness influences the performance of GPT-4o mini, Gemini 2.0 Flash, and Llama4 Scout models.

\subsection{Experiment set-up}

Each question was evaluated under three tonal variants (Neutral, Very Polite, and Very Rude) for all models. GPT-4o mini was accessed through the OpenAI API using the OpenAI Python library (\texttt{openai.AsyncOpenAI}) with model name \texttt{gpt-4o-mini}. Llama-4-Scout was queried via Together AI using the same library (\texttt{openai.AsyncOpenAI}). The API endpoint was set to \url{https://api.together.xyz/v1}, and the model name was \texttt{meta-llama/Llama-4-Scout-17B-16E-Instruct}. 
Gemini~2.0~Flash was accessed through the \texttt{google.generativeai} library with model name \texttt{models/gemini-2.0-flash}.

Each prompt was evaluated over 10 independent runs to mitigate potential stochastic variation in model responses. For each model and each tone condition, we computed Accuracy, defined as the proportion of correctly answered questions, averaged across the 10 runs for all questions within each task. We then compared performance differences across tones and across models using Accuracy as the primary evaluation metric, accompanied by 95\% confidence interval estimates to quantify uncertainty \cite{vv7}.

\subsection{Evaluation Metrics}
To rigorously evaluate the effect of interaction tone on model performance, we analyze mean differences in accuracy between politeness levels (Very Polite vs. Neutral, Neutral vs. Very Rude, and Very Polite vs. Very Rude) across all six question domains and three LLMs. To evaluate the statistical significance of these estimates, we accompany each mean difference with a 95\% confidence interval. The confidence interval does not represent a performance metric but rather quantifies the uncertainty associated with the observed difference, indicating whether it is likely to persist under repeated sampling.

\vspace{1em}
\subsubsection{Mean Difference in Accuracy}
For each pairwise tone comparison, the mean difference measures the average shift in accuracy between two tones. Let $D_{i}^{(t)}$ denote the accuracy of question $i$ under tone condition $t$. Given two tone conditions $t_1$ and $t_2$, the mean difference is computed as:
\begin{equation}
    \Delta = \frac{1}{N} \sum_{i=1}^{N} \left( D_{i}^{(t_1)} - D_{i}^{(t_2)} \right),
\end{equation}
where $N$ is the number of questions in the domain. A positive value of $\Delta$ indicates that the model performs better under tone $t_1$ than $t_2$, whereas a negative value suggests the opposite. Since each question's accuracy is averaged over 10 runs, the mean difference reflects a stable estimate of the tone-induced performance shift rather than run-level noise. Furthermore, because each question is evaluated under both tone conditions, comparisons are performed using paired differences at the question level. This paired design controls for inherent variation in question difficulty and isolates the effect of tone on model performance.

\vspace{1em}
\subsubsection{95\% Confidence Interval}
To evaluate whether the observed mean difference reflects a consistent trend rather than sampling variance, we compute a 95\% confidence interval for the mean difference. The confidence interval provides an estimated range in which the mean difference would lie if future experiments were repeated under the same conditions.

Given the sample of paired accuracy differences $d_i$, the 95\% confidence interval is computed as:

\begin{equation}
\Delta \pm t_{0.975,\,N-1} \cdot \frac{s}{\sqrt{N}}
\end{equation}

where $s$ is the sample standard deviation of the paired differences and $z_{0.975}$ is the critical value for the 95\% confidence level. When the resulting interval does not include zero, the difference between tones is considered statistically significant at the 5\% significance level. On the other hand, when the zero lies within the range, the data do not support a meaningful difference in model accuracy between the two tones.

\vspace{1em}
\subsubsection{Interpretation in Pairwise Tone Comparison}
Together, the mean difference and the 95\% confidence interval provide complementary insights. The mean difference quantifies the direction and magnitude of politeness tone effects, while the confidence interval provides a principled estimate of the statistical reliability of those effects. Taken together, these measures enable robust assessment of whether politeness levels systematically influence LLM model accuracy across domains.

Because multiple pairwise tone comparisons are conducted, the results should be interpreted with awareness of potential multiple-comparison effects. However, since the tone comparisons were predefined and limited in number, we report unadjusted confidence intervals and focus on consistent patterns observed across domains and models.

Table~\ref{tab:model_tone_accuracies} and Table~\ref{tab:cross_domain_tone_grouped} summarize the pairwise accuracy differences across politeness tones for each domain, task, and model. The analysis focuses on two statistical indicators: the mean difference, which captures the directional impact of tone, and the 95\% confidence interval, which characterizes the reliability of the observed differences.

\section{Results}
\begin{table*}[!t]
    \centering
    \scriptsize
    \caption{Task Level Model Accuracy at different tones}
    \label{tab:model_tone_accuracies}
    \resizebox{\textwidth}{!}{%
    \begin{tabular}{lllccccccccc}
        \toprule
        \multirow{2}{*}{Domain} & \multirow{2}{*}{Task} & \multirow{2}{*}{Tone comparison} &
        \multicolumn{3}{c}{GPT-4o-mini} & \multicolumn{3}{c}{Gemini 2.0 Flash} & \multicolumn{3}{c}{Llama4 Scout} \\
        \cmidrule(lr){4-6} \cmidrule(lr){7-9} \cmidrule(lr){10-12}
        & & & Mean diff & SS/NSS & 95\% CI & Mean diff & SS/NSS & 95\% CI & Mean diff & SS/NSS & 95\% CI \\
        \midrule
        \multirow{9}{*}{STEM}
        & \multirow{3}{*}{Anatomy}
        & Very Polite vs Very Rude & +1.23\% & NSS & [-1.33, 5.28] & +0.74\% & NSS & [-1.8, 3.28] & +1.73\% & NSS & [-3.25, 3.75] \\
        & & Neutral vs Very Rude      & +1.73\% & NSS & [-1.83, 5.28] & +0.74\% & NSS & [-2.16, 3.64] & +0.25\% & NSS & [-3.77, 4.26] \\
        & & Very Polite vs Neutral  & +0.25\% & NSS & [-3.39, 3.88] & 0\%     & NSS & [-1.39, 1.39] & 0\%     & NSS & [-1.83, 1.83] \\
        & \multirow{3}{*}{Astronomy}
        & Very Polite vs Very Rude & +0.88\% & NSS & [-0.49, 2.24] & +1.97\% & NSS & [-0.86, 4.81] & -1.10\% & NSS & [-3.73, 1.54] \\
        & & Neutral vs Very Rude      & +1.10\% & NSS & [-1.39, 3.59] & +1.75\% & NSS & [-1.12, 4.62] & +0.66\% & NSS & [-2.26, 3.57] \\
        & & Very Polite vs Neutral  & +0.22\% & NSS & [-2.48, 2.04] & +0.22\% & NSS & [-1.77, 2.21] & -1.75\% & NSS & [-3.77, 0.27] \\
        & \multirow{3}{*}{College Biology}
        & Very Polite vs Very Rude & +1.39\% & NSS & [-0.55, 3.32] & +0.69\% & NSS & [-0.68, 2.07] & +2.08\% & NSS & [-0.28, 4.44] \\
        & & Neutral vs Very Rude      & +1.39\% & NSS & [-1.2, 3.98]  & 0\%     & NSS & [-1.95, 1.95] & +1.39\% & NSS & [-1.36, 4.13] \\
        & & Very Polite vs Neutral  & 0\%     & NSS & [-1.72, 1.72] & +0.69\% & NSS & [-1.69, 3.08] & +0.64\% & NSS & [-3.08, 1.69] \\
        \midrule
        \multirow{9}{*}{Humanities}
        & \multirow{3}{*}{US History}
        & Very Polite vs Very Rude & -0.49\% & NSS & [-1.82, 0.84] & +0.49\% & NSS & [-0.48, 1.46] & +0.82\% & NSS & [-0.98, 2.61] \\
        & & Neutral vs Very Rude      & +0.65\% & NSS & [-1.23, 2.53] & 0\%     & NSS & [-1.37, 1.37] & 0\%     & NSS & [-1.37, 1.37] \\
        & & Very Polite vs Neutral  & -1.14\% & NSS & [-3.1, 0.81]  & +0.49\% & NSS & [-0.48, 1.46] & +0.82\% & NSS & [-0.98, 2.61] \\
        & \multirow{3}{*}{Philosophy}
        & Very Polite vs Very Rude & +0.96\% & NSS & [-0.98, 2.91] & -0.64\% & NSS & [-2.6, 1.31]  & +1.82\% & NSS & [-0.49, 4.14] \\
        & & Neutral vs Very Rude      & +3.11\% & SS  & [0.81, 5.41]  & -0.42\% & NSS & [-2.72, 1.87] & +3.22\% & SS  & [0.87, 5.56] \\
        & & Very Polite vs Neutral  & -2.14\% & SS  & [-3.77, -0.51] & -0.21\% & NSS & [-2.06, 1.63] & -1.39\% & NSS & [-3.36, 0.57] \\
        & \multirow{3}{*}{Professional Law}
        & Very Polite vs Very Rude & +0.67\% & NSS & [-0.74, 2.08] & +0.8\%  & NSS & [-1.16, 2.76] & +1.47\% & NSS & [-0.2, 3.13] \\
        & & Neutral vs Very Rude      & 0\%     & NSS & [-1.32, 1.32] & -0.67\% & NSS & [-2.66, 1.33] & +1.93\% & SS  & [0.06, 3.8] \\
        & & Very Polite vs Neutral  & +0.67\% & NSS & [-1.07, 2.41] & +1.47\% & NSS & [-0.53, 3.46] & -0.47\% & NSS & [-2.15, 1.22] \\
        \bottomrule
    \end{tabular}
    }

    \vspace{0.5em}
    \footnotesize Mean diff is reported in percentage points. SS = statistically significant; NSS = not statistically significant.
\end{table*}

\begin{table*}[!t]
    \centering
    \scriptsize
    \caption{Domain Level Model Accuracy at different tones}
    \label{tab:cross_domain_tone_grouped}
    \resizebox{\textwidth}{!}{%
    \begin{tabular}{lccccccccc}
        \toprule
        \multirow{2}{*}{Tone comparison} &
        \multicolumn{3}{c}{GPT-4o-mini} &
        \multicolumn{3}{c}{Gemini 2.0 Flash} &
        \multicolumn{3}{c}{Llama4 Scout} \\
        \cmidrule(lr){2-4} \cmidrule(lr){5-7} \cmidrule(lr){8-10}
        & Mean diff & SS/NSS & 95\% CI & Mean diff & SS/NSS & 95\% CI & Mean diff & SS/NSS & 95\% CI \\
        \midrule
        \multicolumn{10}{l}{\textbf{STEM}} \\
        Very Polite vs Very Rude & +1.39\% & SS  & [0.09, 2.69]  & +1.16\% & NSS & [-0.19, 2.51] & +0.39\% & NSS & [-1.24, 2.02] \\
        Neutral vs Very Rude       & +1.38\% & NSS & [-0.26, 3.04] & +0.85\% & NSS & [-0.65, 2.35] & +0.77\% & NSS & [-1.08, 2.62] \\
        Very Polite vs Neutral   & +0.02\% & NSS & [-1.49, 1.49] & +0.31\% & NSS & [-0.83, 1.45] & -0.39\% & NSS & [-0.82, 1.59] \\
        \addlinespace
        \multicolumn{10}{l}{\textbf{Humanities}} \\
        Very Polite vs Very Rude & +0.53\% & NSS & [-0.50, 1.55] & +0.30\% & NSS & [-0.86, 1.45] & +1.44\% & SS  & [0.31, 2.58] \\
        Neutral vs Very Rude       & +1.08\% & NSS & [-0.12, 2.29] & -0.46\% & NSS & [-1.70, 0.78] & +1.94\% & SS  & [0.74, 3.14] \\
        Very Polite vs Neutral   & -0.56\% & NSS & [-1.67, 0.56] & +0.76\% & NSS & [-0.39, 1.90] & -0.49\% & NSS & [-1.58, 0.59] \\
        \bottomrule
    \end{tabular}
    }
    \vspace{0.5em}
    \footnotesize Mean diff is reported in percentage points. SS = statistically significant; NSS = not statistically significant.
\end{table*}

Across both STEM and Humanities domains, the direction of the mean differences consistently favors more neutral or polite tones over rude tones. Positive mean differences appear in the large majority of tone comparisons: 27 out of 36 model–task comparisons involving Very Rude prompts show higher accuracy under Neutral or Very Polite tones. Conversely, cases where Very Rude prompts outperform more polite tones are uncommon and small in magnitude, with only 5 out of 36 comparisons exhibiting negative mean differences. This directional pattern indicates that impolite phrasing rarely improves accuracy for the three evaluated LLMs and generally leads to marginally worse performance.

\subsection{Statistical Significance by Task and Model}
\vspace{0.5em}
Although directional differences are observed across most tasks, the majority of 95\% confidence intervals include zero, suggesting non–statistically significant (NSS) outcomes for most tone comparisons. Nonetheless, several statistically significant (SS) effects do emerge, and all of them occur within the Humanities domain.

\begin{itemize}
\vspace{0.5em}
\item \textbf{Philosophy}: 
Statistically significant tone effects are observed in both the GPT and Llama models. For GPT, significant accuracy differences is observed in the comparisons of \textit{Neutral vs.\ Very Rude} and \textit{Very Polite vs.\ Neutral}. The mean difference for the Neutral–Very Rude comparison is +3.11\%, indicating that neutral prompts tend to produce higher accuracy than Very Rude prompts. In contrast, the Very Polite–Neutral comparison yields a mean difference of -2.15\%, suggesting that Very Polite prompts result in lower accuracy than neutral prompts. This pattern indicates that, for philosophical questions, increasingly politeness in phrasing does not necessarily improve GPT's performance. For Llama, a +3.22\% statistically significant accuracy difference is observed when comparing the Neutral v.s. Very Rude tones. Taken together, both GPT and Llama consistently show that Very Rude prompts lead to reduced accuracy in the Philosophy task.

\vspace{0.5em}
\item \textbf{Professional Law}: Statistically significant tone effect is observed in the Llama model for the \textit{Neutral vs.\ Very Rude} comparison. The positive mean difference of +1.93\% indicates that neutral prompts yield higher accuracy than Very Rude prompts. This finding mirrors the pattern observed in the Philosophy task, reinforcing the conclusion that rude tone prompts negatively affect Llama's performance on Humanities tasks.
\end{itemize}

\vspace{0.5em}
In contrast, the Gemini model shows no statistically significant tone effects across all evaluated tasks. This consistent NSS pattern suggests that Gemini's accuracy is comparatively stable under tone variation. Although the small directional differences in mean accuracy suggests that there may be similar tonal outcome differences for Gemini, we didn't observe conclusive differences from our experiment.

\subsection{Model-Specific Sensitivity to Tone}
\vspace{0.5em}
When viewed as a whole, the results suggest that GPT and Llama exhibit measurable tone sensitivity, particularly within Humanities tasks that involve higher abstraction or more nuanced reasoning. These effects are not only directionally consistent but occasionally statistically significant. In comparison, the Gemini model demonstrates minimal sensitivity, showing no significant accuracy differences across tone conditions. This difference across model families implies that the architecture or training differences behind each model may modulate how models respond to different prompts phrasings.

\subsection{Humanities vs.\ STEM Domains}
\vspace{0.5em}
Tone effects tend to be more pronounced in Humanities tasks, which involve higher-level reasoning, nuance, and interpretive judgment. In contrast, STEM tasks show consistently positive but statistically weaker effects, with most confidence intervals crossing zero. This suggests that, despite modest gains from more polite or neutral tones in STEM, the question-level variability often prevents effect sizes from achieving statistical significance. These domain-specific patterns align with prior findings that model responsiveness to politeness varies by subject domain and contextual depth ~\cite{lj9}.

\subsection{Aggregated Domain Level Effects}

\vspace{0.5em}
To evaluate whether tone effects persist when users ask questions across diverse subject areas, we aggregated question-level differences across the three tasks within each domain and recomputed the statistical tests. At the aggregated domain level, the directional trends observed in mean differences remain consistent with task-level observations; however, most confidence intervals include zero, indicating predominantly NSS outcomes.

\vspace{0.5em}
This suggests that when users engage models on a broad range of topics, the probability that prompt tone materially alters overall accuracy becomes small. Tone-induced variability observed at the per-task level tends to attenuate when aggregated, indicating that tone effects, while occasionally present in specific contexts, do not systematically degrade accuracy across general-purpose usage scenarios.

\subsection{Overall Interpretation}
Taken together, the results indicate that:
\begin{itemize}
\item Very Polite or Neutral tones tend to yield higher accuracy than Very Rude tones across most tasks.
\item Very Polite tone does not always yield better model performances than Neutral tone.
\item Statistically significant tone effects are rare and concentrated in Humanities tasks for the GPT and Llama models.
\item Gemini shows no significant tone sensitivity.
\item When questions are aggregated across domains, tone effects diminish and become negligible.
\end{itemize}

\vspace{0.5em}
These findings suggest that tone can influence model performance in certain tasks—particularly within interpretive or linguistically nuanced domains—but its impact diminishes in broad mixed-domain usage, supporting the robustness of modern LLMs under typical user interaction conditions.

\section{Conclusion}
This study examined how variations in interaction, from Very Polite to Very Rude, affect the accuracy of three contemporary large language models across six representative MMMLU tasks spanning STEM and Humanities domains. Using repeated trials and statistical evaluation via pairwise mean differences and 95\% confidence intervals, we provide an empirical characterization of tone sensitivity for GPT-4o-mini, Gemini 2.0 Flash, and Llama4 Scout.

\vspace{0.5em}
Overall, tone shows small directional effects on model performance, though most comparisons do not reach statistical significance. Neutral and Very Polite prompts generally yield higher accuracy than Very Rude prompts, yet statistically significant effects arise only in a small subset of Humanities tasks, particularly Philosophy and Professional Law. GPT and Llama exhibit noticeable tone sensitivity in these areas, whereas Gemini shows no statistically significant effects across any tone comparison. When aggregating performance across domains, tone effects diminish substantially for all models, suggesting that tonal variation is unlikely to materially impact accuracy in broad, mixed-domain usage scenarios.

\vspace{0.5em}
These results complement and extend prior work on prompt politeness. Earlier studies based on substantially smaller question sets reported different directional conclusions regarding the benefit of impolite prompts, underscoring the importance of dataset scale and representativeness in evaluating tone effects. Our findings suggest that, while tone may influence performance in specialized, interpretive tasks, modern LLMs demonstrate strong robustness to tonal variation in typical real-world interactions.

\vspace{0.5em}
Future work may extend this analysis along several directions. First, evaluating a broader set of models, including fully open-source architectures with transparent parameter counts, training corpora, and routing mechanisms, would help clarify whether tone robustness is shaped more by architectural design, data curation differences, or instruction-tuning strategies. Second, moving beyond English multiple-choice benchmarks to multilingual, open-ended, and multimodal tasks could reveal stronger or qualitatively different tone effects in more naturalistic interaction settings. Third, using larger and more domain-representative evaluation datasets would help further generalize the findings; prior work based on only fifty questions \cite{lj10} reported trends that differ substantially from those observed here, suggesting that dataset scale and coverage materially influence the detection of tone effects. Finally, exploring richer tone manipulations (e.g., degrees of formality, affect, or emotional intensity) and incorporating additional evaluation metrics such as calibration error, safety compliance, and user-perceived helpfulness would provide a more comprehensive understanding of how interaction tone influences the reliability and usability of LLMs in practical deployments.

\end{document}